# Residential Construction Image Dataset Using CSI Divisions for Deep Learning-based Object Detection

**Fahad Iqbal[1], and Shayan Mirzabeigi[1,2]**

[1] Department of Sustainable Resources Management, State University of New York College of Environmental Science and Forestry, Syracuse, NY 13210, USA

[2] Department of Mechanical and Aerospace Engineering, Syracuse University, Syracuse, NY 13244, USA

Corresponding author: Shayan Mirzabeigi (smirzabeigi@esf.edu)

## Abstract

Accurate recognition of construction site elements is essential for the application of deep learning (DL) in the construction industry. Despite rapid advances in computer vision, the adoption of DL in construction has been limited by the lack of structured, publicly available datasets aligned with established industry standards. To address this gap, we introduce the Residential Construction Dataset Using CSI Divisions (ResCD-CSI) image dataset, a manually labeled image benchmark organized according to the divisional structure defined by the Construction Specifications Institute (CSI). The dataset comprises high-resolution images collected from active residential construction sites across the United States, capturing diverse building components, materials, and equipment under real-world conditions. In total, 16,365 objects spanning 14 major CSI divisions were precisely annotated using bounding boxes, ensuring consistency and reliability for object detection tasks. By aligning with the CSI divisional framework, ResCD-CSI establishes a standardized resource that bridges construction practice with computer vision research and supports the development of Artificial Intelligence-driven solutions for residential construction monitoring.

## Background & Summary

The construction industry remains highly labor-intensive, particularly for skilled labor, with most on-site management tasks and interventions still relying on manual judgments, which are often inefficient and difficult to scale [1]. Although monitoring technologies are increasingly deployed to capture construction activities, extracting and interpreting real-time site information still requires considerable human effort. In this context, the rapid advancement of computer vision offers a transformative opportunity for transition to a predictive construction management approach. By automating tasks traditionally dependent on human oversight, computer vision technologies can significantly enhance safety management, progress monitoring and tracking, and operational efficiency [2]. This potential has led both academia and industry to increasingly recognize the role of vision-based techniques in construction, with early applications ranging from detecting unsafe worker behaviors to identifying violations such as missing personal protective equipment (PPE) and automating progress monitoring [3].

In addition, deep learning (DL)-based object detection has achieved high precision in controlled benchmarks, though performance often degrades in complex, unstructured environments like construction sites. Compared to traditional image processing methods, DL offers superior speed, robustness, and automated feature extraction [4]. These advantages make DL object detection a promising direction for construction site management. However, its practical application in this field remains constrained by a fundamental limitation, including the lack of large-scale, well-annotated, and publicly available datasets tailored to the complexity of construction environments [5]. Existing datasets are often small and narrowly focused on specific objects, such as PPE, workers, or machinery, and rarely capture the broader scope of construction activities. Several challenges drive these gaps: (1) restricted access to active sites due to safety and security concerns; (2) limited and repetitive camera perspectives that risk overfitting detection models; and (3) the inherently dynamic and chaotic nature of construction environments, which increases the difficulty and cost of data collection while demanding expert annotation [6].

In recent years, multiple datasets have been introduced to address specific aspects of construction. Examples include worker and PPE-focused collections such as GDUT-Hardhat Wearing Detection[7], Hard Hat Workers[8], equipment-centered datasets such as the Alberta Construction Image dataset[9] and the Advanced Infrastructure Management Group dataset[10], and broader resources such as Pictor-v2/v3[11,12] and SODA [13]. Specialized datasets targeting the rebar Steel Bar dataset[14], Synthetic Rebar dataset[15] and safety harness detection[16] have also been published. While valuable, these datasets are often self-established, limited in scale, and focused on narrow categories. A recent review highlighted that over half (53.3%) of publicly available datasets for the construction industry concentrate on PPE detection, 20% on equipment, and only 26.6% extend to other categories, with significant contributions emerging from China, while public availability remains limited [17].

To advance object detection research in the construction industry, a large-scale image dataset containing key site objects (e.g., workers, materials, machinery, and layouts) is essential. A comprehensive dataset would not only support the development of DL–based object detection models but also provide a reliable benchmark for evaluating algorithm performance.[18] Given the professional expertise required to create such a dataset, as well as the challenge of data availability, its development remains an open challenge. Moreover, existing datasets are typically limited to infrastructure and commercial projects, leaving residential construction underrepresented despite its economic significance and distinct site characteristics. Furthermore, existing datasets often adopt inconsistent category definitions, limiting their applicability across the industry. A standardized dataset organized according to an established construction divisions classification, e.g., Construction Specifications Institute (CSI), could provide a consistent framework, enhancing comparability and improving adoption in both research and practice.

Further reinforcing this gap, recent studies have emphasized the need for more comprehensive datasets. For instance, Zhang et al. explored vision-based tracking of concrete pouring and recommended developing complete datasets [19]. Hou et al. applied computer vision for safety risk analysis and highlighted the role of advanced models [20]. Duan et al. introduced the SODA dataset but pointed to the need for richer annotations [13]. Kanghyeok et al. developed a small tool database for safety monitoring, suggesting broader model applications[21], while Ekanayake et al. reviewed interior progress monitoring research and identified the lack of large, diverse datasets [22].

To address these gaps, the objective of this study is to develop a publicly available residential construction image dataset that overcomes these limitations. The developed dataset is organized according to the divisional structure defined by the CSI and includes high-resolution images collected from active residential sites across the United States. A total of 16,365 objects spanning 14 major divisions were annotated with bounding boxes. The dataset also enables categorization by construction stages (e.g., preconstruction, construction, rough-in/services, finishing), offering structured annotations that can support diverse research and applied use cases. By providing a broader and standardized dataset aligned with industry practices, this resource supports object detection research in construction and creates opportunities for reuse in applications such as safety assessment, progress tracking, and automated monitoring.

# Methods

The Residential Construction Dataset Using CSI Divisions (ResCD-CSI) dataset was developed through a structured process designed to ensure both industry relevance and research utility. Compared to conventional approaches that rely heavily on online resources, the images in this dataset were primarily acquired on actual construction sites, providing clearer contextual relevance for object detection and construction monitoring tasks. The development process of this dataset mainly included four steps: CSI categories selection, image acquisition, data cleaning, and image annotation, as shown in Figure 1.

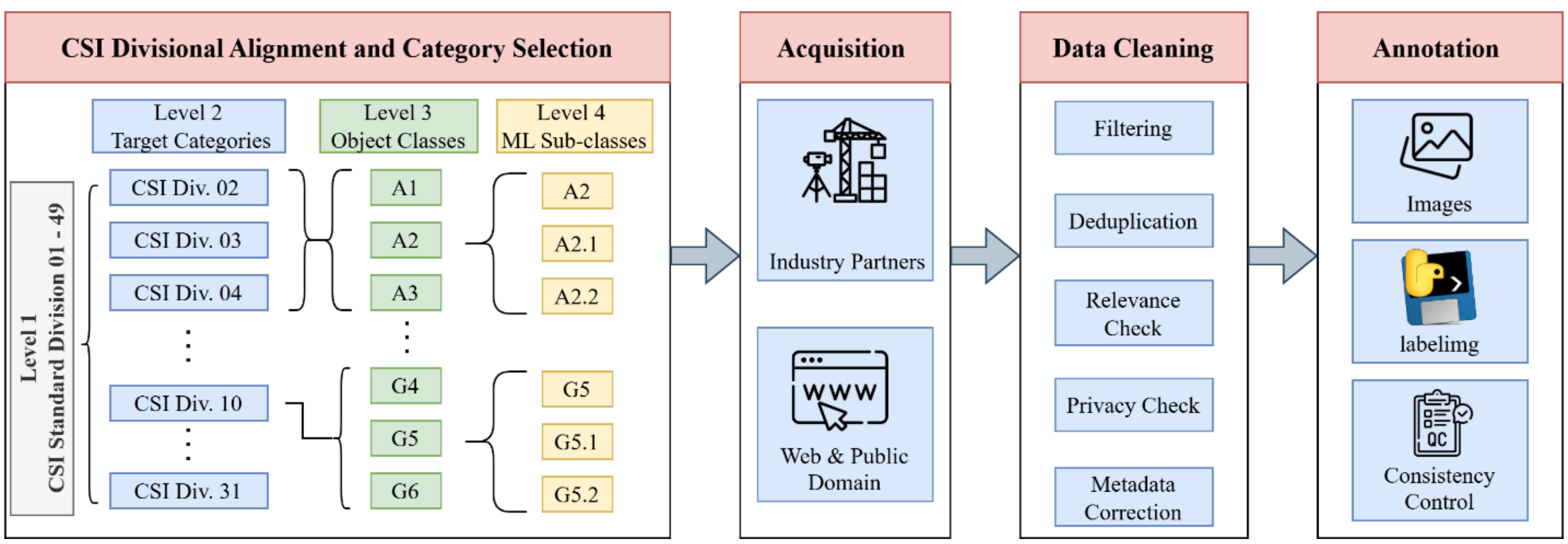

Figure 1. Overview of the development process for the ResCD-CSI dataset.

**CSI category selection.** In previous studies, researchers have often applied their own classification schemes for construction site images. For example, Zhou et al.[23] used the 4M1E principle, which divides construction risk factors into man, material, machine, method, and environment. Based on this principle, visible objects were grouped into four categories: worker, material, machine, and layout. Such classifications have proven useful for safety analysis and risk management in construction. In this study, however, object categories were defined according to the CSI MasterFormat divisions. The CSI-based system was selected because it is widely used in construction project management for cost estimation, scheduling, monitoring, and documentation, and therefore provides a direct link between computer vision datasets and established industry practices. While the 4M1E framework offers a broader conceptual view of construction activities, the CSI divisions allow a more practical and standardized categorization that aligns with project-level decision-making. A list of the CSI divisions targeted in this dataset is provided in Table 1.

The selection process of these divisions followed a structured hierarchy to bridge industry standards with computer vision requirements. Level 1 represents the standard CSI MasterFormat (Divisions 01–49). From these, Level 2 Target Categories were identified by selecting only the divisions that manifest as physical elements on residential sites, excluding abstract or administrative divisions. Level 3 defines the specific Object Classes, such as A1 through A5, which represent distinct physical entities that can be isolated and annotated within an image. For certain categories, an additional Level 4 of ML Sub-classes was implemented to manage environmental and visual variations. A primary example is the division of Windows into *'Windows_Indoor'* and *'Windows_Outdoor'* to handle different environmental contexts and improve detection reliability. To ensure a robust ground truth, specific divisions were merged into these classes based on their functional and visual overlap. For instance, the *'A'* section, which includes *'ExcavationSoil'* through *'PrecastConcreteBlock'*, collectively maps to Divisions 02, 03, and 31. This multi-level alignment ensures that while the dataset is optimized for machine learning, it remains fully traceable to construction industry standards. A visual description of each class can also be seen in Figure 2, while Figure 3 shows a histogram of classes indicating the number of images and objects per class.

Table 1. Descriptions of classes and corresponding Construction Specifications Institute (CSI) divisions for the dataset.

| **S. No** | **Alias** | **Class** | **CSI Divisions** | **Description** |
|---|---|---|---|---|
| 0 | A1 | ExcavationSoil | 02, 03, 31 | Dirt work, grading, trenching, site prep |
| 1 | A2 | Formwork | | Concrete forms, molds, reusable systems |
| 2 | A3 | RebarReinforcement | | Steel bars, mesh, wire ties |
| 3 | A4 | PouredConcreteSlabs | | Footings, foundations, floor slabs |
| 4 | A5 | PrecastConcreteBlock | | Factory-made panels, beams, slabs |
| 5 | B1 | WoodFraming | 04, 05, 06 | Studs, joists, rafters, trusses |
| 6 | B2 | SteelFraming | | Beams, columns, connectors |
| 7 | B3 | Sheathing&Bracing | | Plywood, OSB, bracing panels |
| 8 | B4 | RoofFraming | | Trusses, rafters, purlins |
| 9 | C1 | ExteriorWalls | 07, 08 | Brick, siding, stone veneer |
| 10 | C2 | RoofingSystems | | Shingles, metal roofing, membranes |
| 11 | C3 | Windows | | Casement, sliding, double-pane |
| 12 | C4 | Doors | | Entry, garage, interior doors |
| 13 | C5 | Waterproofing&Insulation | | Spray foam, barriers, boards, batts |
| 14 | D1 | DrywallFraming | 09 | Gypsum boards, finishes, framing |
| 15 | D2 | Ceilings | | Drop ceilings, drywall, tiles |
| 16 | D3 | Flooring | | Tile, wood, carpet, laminate |
| 17 | D4 | WallFinishes | | Paint, wallpaper, paneling |
| 18 | E1 | PlumbingRough-ins | 22 & 23 | Pipes, drains, PEX, copper |
| 19 | E2 | PlumbingFixtures | | Sinks, toilets, showers, tubs |
| 20 | E3 | HVACEquipment | | Ducts, vents, heat pumps, AC |
| 21 | E4 | MechanicalPiping&Insulation | | Piping and insulation systems |
| 22 | F1 | Conduit&Wiring | 26 | Conduit, cable, junction boxes |
| 23 | F2 | ElectricalPanels&Breakers | | Panels, breakers, fuses |
| 24 | F3 | LightingFixtures | | Recessed, pendant, outdoor |
| 25 | F4 | SolarPanels&Batteries | | PV panels, inverters, batteries |
| 26 | G1 | Cabinetry&Millwork | 10 | Cabinets, vanities, shelving |
| 27 | G2 | Fireplaces&Chimneys | | Fireplaces, chimneys, flues |
| 28 | G3 | BathroomAccessories | | Mirrors, grab bars, shelves |
| 29 | G4 | ExteriorDecks&Railings | | Decking, railings, stairs |
| 30 | G5 | ConstructionEquipmentX | | Construction equipment |
| 31 | G6 | ConstructionStaffX | | Construction workforce |
| 32 | B5 | MassTimber | 06 | CLT panels, glulam beams, columns, and heavy timber framing. |

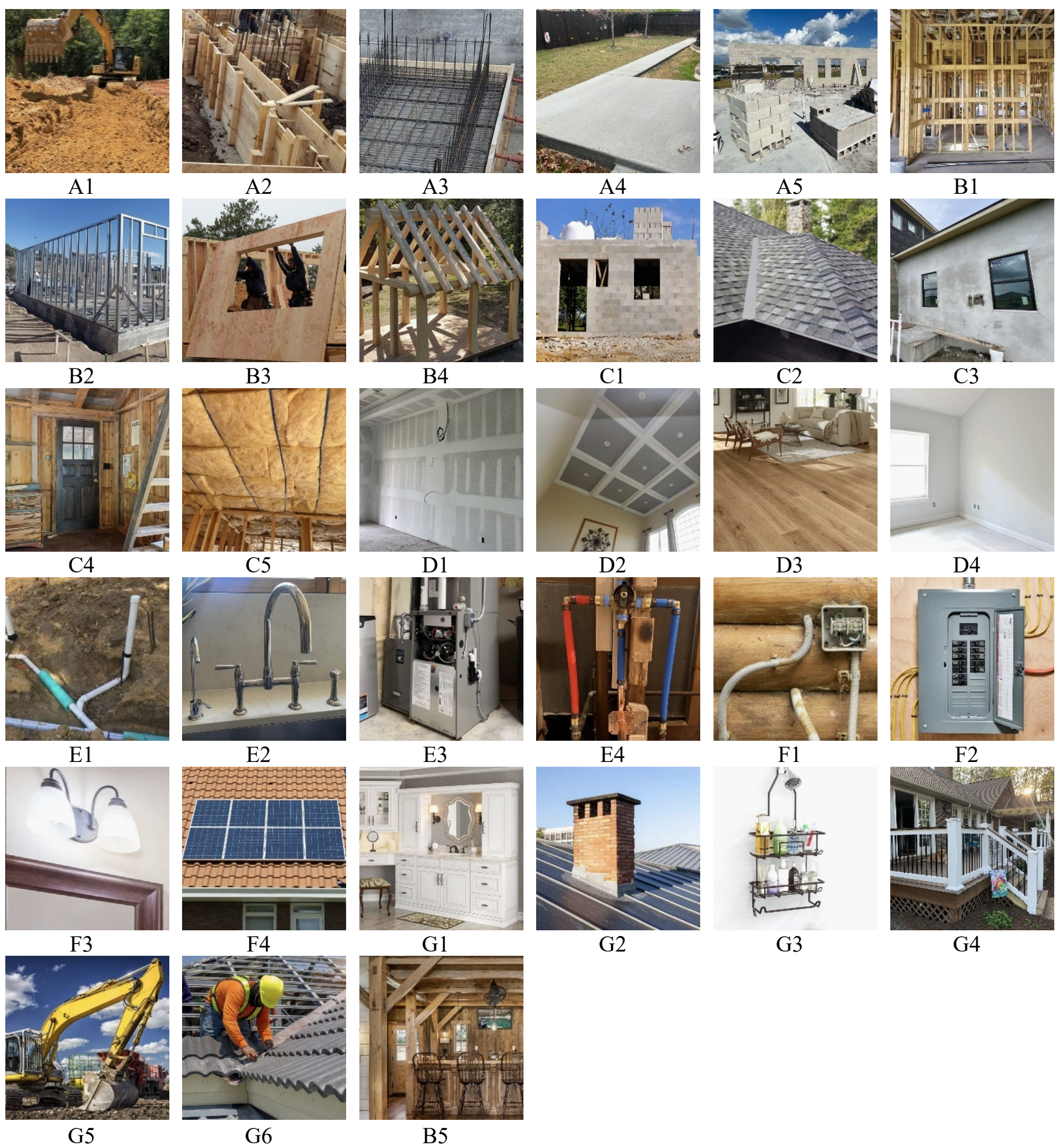

Figure 2. Visual overview of classes (refer to Table 1 for descriptions).

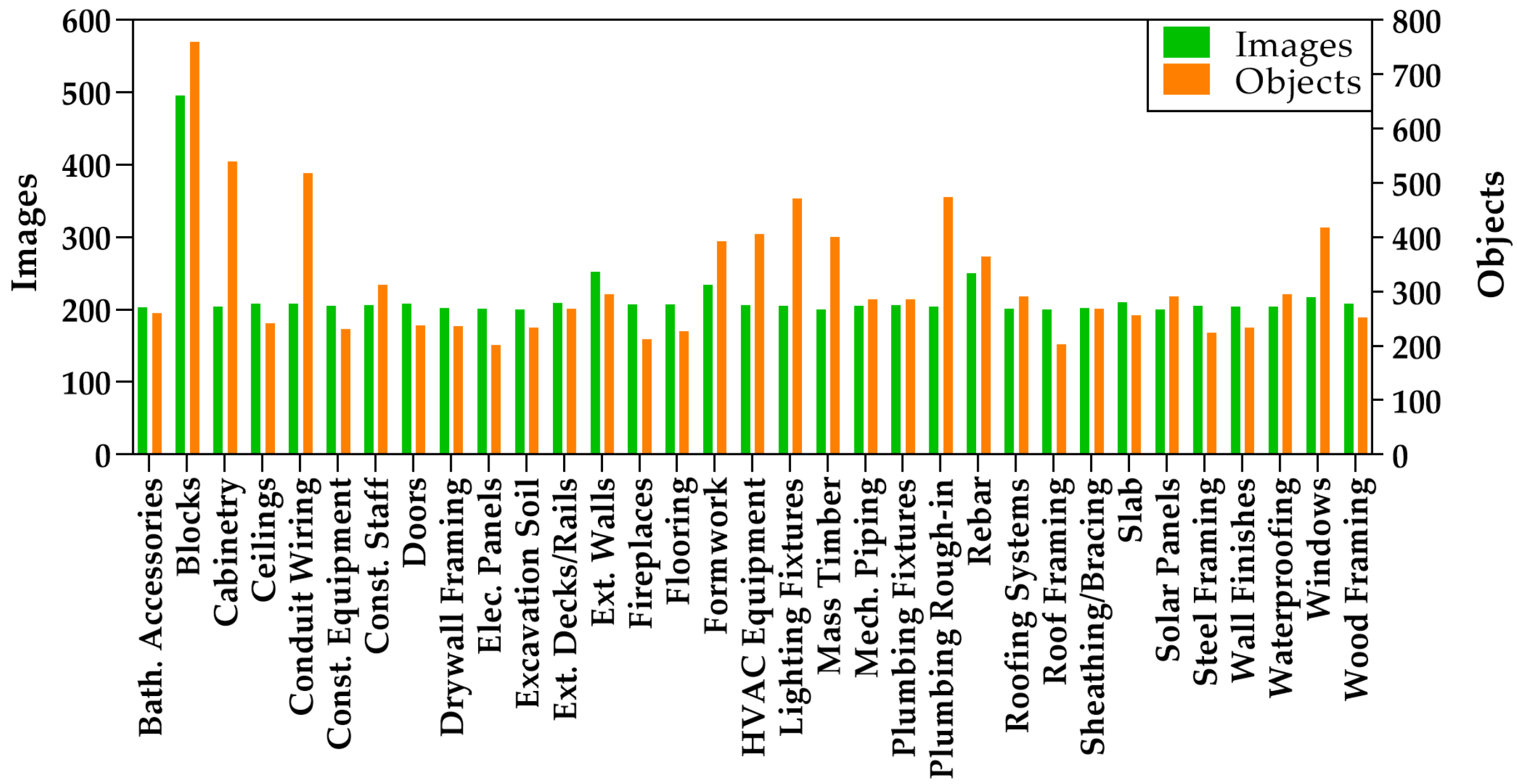

Figure 3. Class-wise distribution of images and objects.

**Data source.** This section is divided into two sub-sections that describe image collection and selection.

*Image collection.* The dataset consists of 9,100 images sourced from both active construction sites and the public web domain. To ensure a high-fidelity, site-specific context, roughly 75% of the data (6,825 images) was pulled directly from the as-built records of 12 different projects across Florida and New York. These images, captured via mobile devices in PNG, JPG, and HEIC formats, cover a broad spectrum of construction environments, activities, and phases. The remaining 25% (2,275 images) were gathered from web sources to fill gaps in underrepresented categories, such as construction personnel and heavy equipment. This hybrid strategy was intentionally used to enhance data diversity and help the model better recognize variations in material appearance and installation techniques. While web-based images add visual variety, the heavy reliance on field records keeps the dataset focused on the actual day-to-day conditions of residential construction. A key limitation of the current ResCD-CSI dataset is its geographic concentration in Florida and New York. This introduces the risk of domain shift, where models trained on these samples may face reduced accuracy when applied to construction sites in regions with significantly different architectural styles, climatic conditions, or regional material preferences. The dataset is designed for iterative expansion; future versions will incorporate data from additional geographic sectors to further reduce regional bias and increase the statistical diversity of the training samples. To address privacy concerns, all images were processed to remove personally identifiable information. In addition, a fixed-ratio cropping tool was applied to standardize formatting and streamline preprocessing, ensuring that the dataset remains consistent, ethically sound, and ready for large-scale analysis.

*Image selection.* From the raw image pool, images were manually reviewed to eliminate duplicates, irrelevant views, or poor-quality captures. After cleaning the dataset, the total number of quality images was reduced to 8,441. The selection of images was prioritized by those that clearly depict construction activities, materials, or progress stages. The curated dataset aims to balance project diversity, construction stage coverage, and visual clarity.

**Data processing.** This section is divided into image annotation, annotation standard, and data augmentation.

*Image annotation.* Accurate image annotation is essential for reliable model development. In this study, labeling was performed manually using domain expertise to ensure consistency and correctness. An object detection framework was adopted, in which specific construction elements within each image are localized and identified using bounding boxes. This setup is commonly applied in computer vision when the objective is to detect and spatially map material types based on their visual characteristics. Establishing a clearly structured classification dataset is therefore critical for effective training and evaluation.

Object extraction was conducted before isolating the relevant visual content. Bounding box coordinates were used to crop target regions from the original images, and the resulting samples were assigned to their respective class labels. While bounding boxes may not provide the pixel-level precision required for highly detailed sorting tasks, and segmentation-based methods have been proposed for finer analysis, box-level annotation remains appropriate for datasets with a large number of categories. In construction-related applications, this approach offers a practical balance between annotation efficiency and classification requirements.

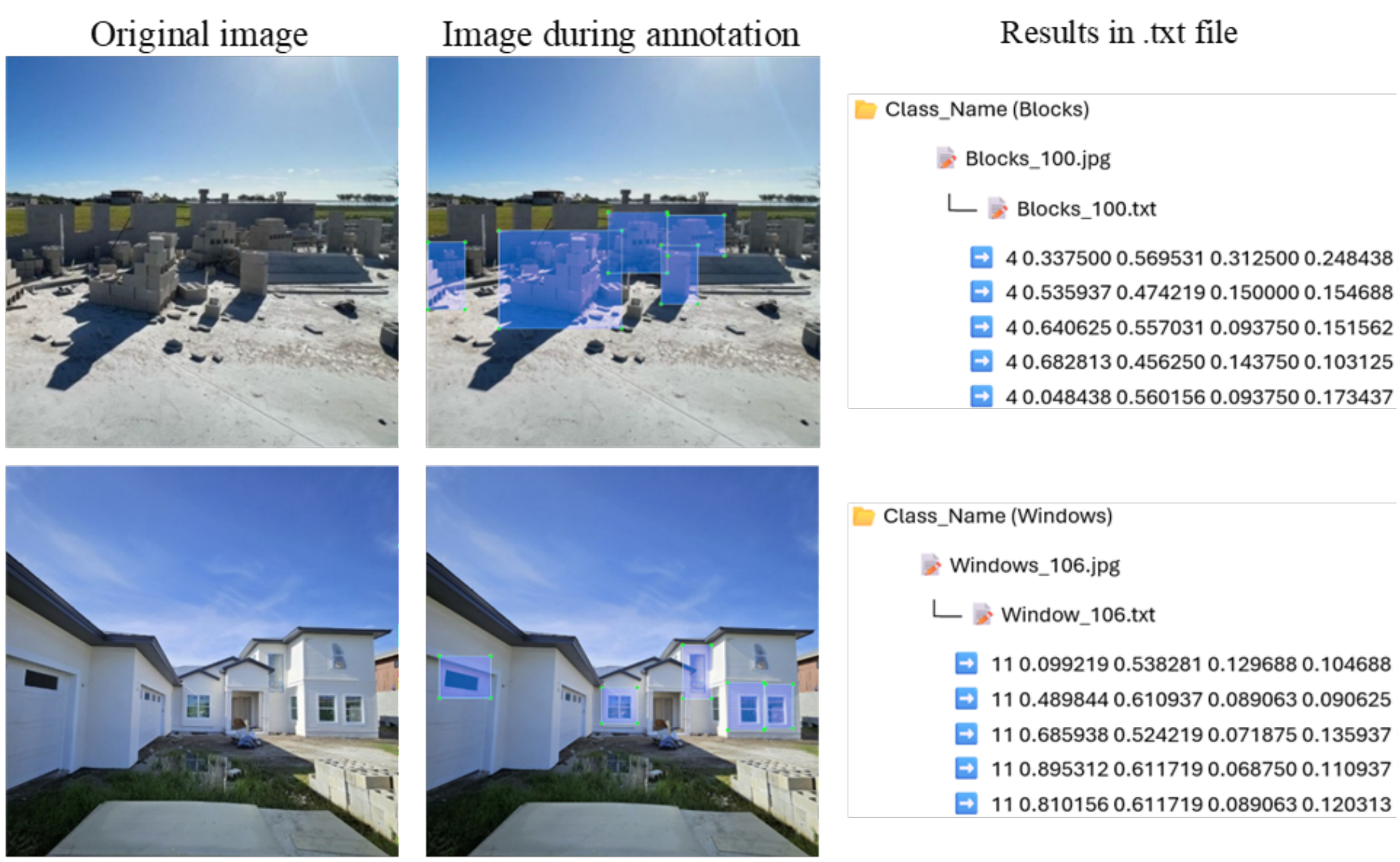

Figure 4. Examples of selected images before, during, and after annotation for object detection.

The annotation process of this dataset was carried out by two undergraduate students and one graduate student, all of whom received prior training in the use of annotation tools and guidelines for applying CSI-based classifications.

*Annotation standard.* Annotations for this dataset followed the predefined CSI categories, marking objects such as structural elements, equipment, and materials. Metadata, including the construction stage, was recorded and linked to each image. Established computer vision guidelines were followed to ensure bounding boxes were consistent in scale, position, and overlap. A structured folder hierarchy and metadata files ensured usability. The You Only Look Once (YOLO) format was adopted as the dataset standard, with each annotation saved in a .txt file containing class ID and normalized bounding box coordinates (x, y, width, height). This format ensures compatibility with widely used object detection frameworks [24]. Figure 4 presents two examples used for manual annotation and detection: the original images, the images during annotation, and the resulting .txt files. The first column shows the raw images in JPG format with a resolution of 640 × 640 pixels. The second column displays the images during labeling and annotation using LabelImg v1.8.6 [25], with annotated instances indicated by blue rectangles.

A systematic quality control process was adopted to ensure the structural integrity and precision of the ResCD-CSI dataset through three primary heuristics. First, Maximum Tightness required bounding boxes to encompass only the visible extent of object pixels, thereby minimizing background noise during feature extraction. Second, Instance Separation mandated that closely spaced or overlapping objects of the same class be annotated as individual instances to support high-density object detection. Finally, Taxonomic Consistency ensured that labels were cross-verified against CSI MasterFormat descriptions, allowing visually similar materials to be correctly categorized based on their functional roles within the residential assembly.

The annotation process encountered several challenges, primarily visual ambiguity and environmental stochasticity, which required a structured validation framework. To resolve identification conflicts between structurally similar classes, such as *Wood_Framing* versus *Roof_Framing* or *Steel_Framing* versus *Conduit_Wiring* as can be seen in Figure 5a, a "context-dominance" rule was applied to categorize objects based on their spatial relationship to construction phases. Intra-class variance from contextual background and fluctuating illumination was mitigated by implementing contextual sub-classes, specifically *Windows_Indoor* and *Windows_Outdoor* as can be seen in Figure 5*b*, to decouple intrinsic morphological features from environmental variables. High-density occlusion in classes like *Rebar_Reinforcement* was managed through a 30% minimum visibility threshold and a "visible-boundary" constraint, ensuring bounding boxes excluded low-information features that could adversely affect the model's loss function. Any inter-annotator discrepancies were resolved through a consensus-based decision to ensure the integrity of the ground truth.

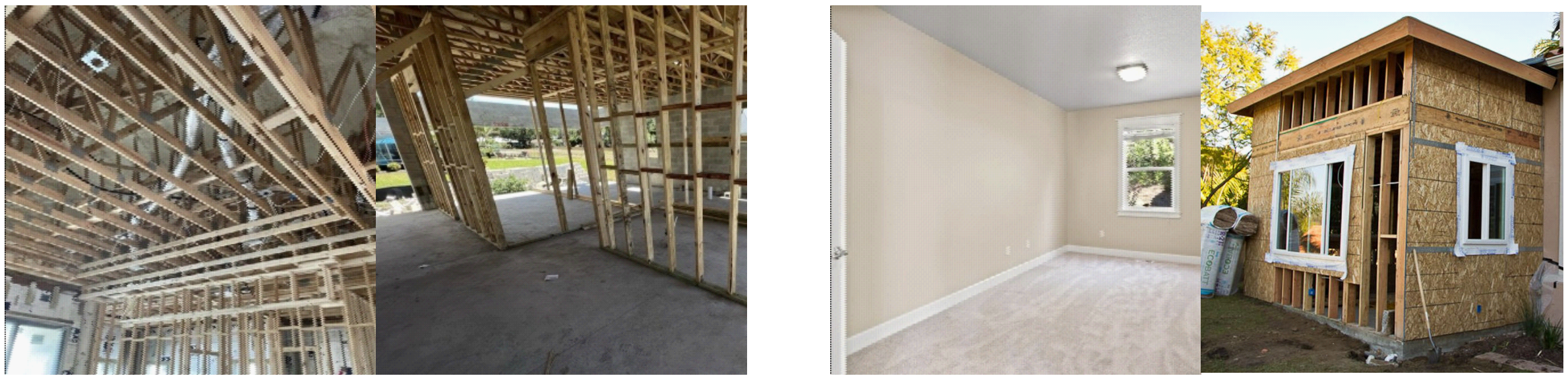

(a) (b)

Figure 5. Annotation examples for challenging cases: (a) Differentiating Wood and Roof Framing, and (b) Indoor vs. Outdoor Window classification.

*Data augmentation.* To enhance dataset robustness and support model generalization, several standard augmentation techniques were applied during the baseline model training phase; however, to maintain the integrity of the ground truth data, no augmented images are included in the published dataset repository. The applied transformations included geometric adjustments such as rotation and horizontal or vertical flipping, along with photometric modifications like noise injection. These techniques simulate the environmental stochasticity of active construction sites, such as varying solar angles and sensor artifacts. Our preliminary results demonstrated that incorporating these augmentations during training increased the baseline model's Mean Average Precision mAP by 5.7%. Data augmentation is widely recognized as a cost-effective strategy to expand dataset diversity and mitigate overfitting in computer vision tasks[26].

# Data Record

The dataset is publicly available on figshare[27] (https://doi.org/10.6084/m9.figshare.30325087) and is organized into three main components: Dataset_Repository.zip, YOLO_Split.zip, Metadata.zip, and a README file. The Dataset_Repository is structured into subfolders by CSI Division, as shown in Figure 6. Within the Class folders, construction site images are stored in JPG format. Each image is paired with a corresponding .txt file containing the annotations. These text files include the class IDs and normalized bounding box coordinates for every object identified in the image. The YOLO_Split folder is for direct use with the notebook file having the same images but divided into train, test, and val folders. The Metadata folder provides additional information on description, construction stage, and image properties, supporting dataset usability and analysis. The README file documents the dataset structure, provides basic instructions for using the data, and includes a change log detailing updates or modifications over time.

```
Dataset_Repository
│
├── Class_Name_01
│   ├── Class_Name_01_001.jpg
│   ├── Class_Name_01_002.jpg
│   ├── Class_Name_01_001.txt
│   └── Class_Name_01_002.txt
│
├── Class_Name_02
│   ├── Class_Name_02_001.jpg
│   ├── Class_Name_02_002.jpg
│   ├── Class_Name_02_001.txt
│   └── Class_Name_02_002.txt
│
├── Class_Name_03
│   ├── Class_Name_03_001.jpg
│   ├── Class_Name_03_002.jpg
│   ├── Class_Name_03_001.txt
│   └── Class_Name_03_002.txt
│
├── Metadata
│   ├── dataset_description.json
│   ├── image_properties.json
│   └── construction_stage.json
│
└── README.md   (includes dataset structure,
     usage notes, and change log)
```

Figure 6. Data repository structure.

## Technical Validation

The validation of the ResCD-CSI dataset focused on the accuracy of the bounding box labels and how well the images represent the complex, real-world conditions of a construction site. To ensure the dataset is ready for practical use, baseline checks were conducted to verify the consistency of labels across various lighting conditions, occlusions, and viewing angles. While the dataset contains naturally occurring class imbalance, as structural elements like wood framing appear more frequently than finishing accessories, this reflects the actual progression of residential projects.

To demonstrate that the annotations are suitable for deep learning, we conducted illustrative tests and validation using the YOLOv11m architecture. This one-stage detector was selected for its balance of speed and accuracy, which is ideal for construction monitoring tasks. All the experiments with the dataset were implemented in PyTorch on Google Colab, utilizing an NVIDIA Tesla T4 GPU. Dataset checks focused on the consistency of bounding box labels, diversity of object categories, and representativeness of real-world construction site conditions, including variations in lighting, occlusions, and viewing angles. To ensure practical usability, baseline object detection models were trained and evaluated on the dataset. Performance was assessed using standard metrics, including mean Average Precision (mAP), precision, and recall, thereby confirming the dataset's reliability for DL–based construction object detection research [32]. Intersection over Union (IoU) was used to quantify the overlap between predicted bounding boxes and ground truth boxes. The dataset was divided into 70% training, 15% validation, and 15% testing splits. Model training was conducted using the YOLOv11m architecture with pretrained weights (yolo11m.pt) for 60 epochs at an input resolution of 640 × 640 pixels. Default YOLO hyperparameters were employed, including a learning rate of 0.001, momentum of 0.937, and weight decay of 0.0005. Early stopping was used to prevent overfitting.

While the YOLOv11m architecture provides a robust and accessible baseline for initial evaluation, we acknowledge that limiting the validation to a single detector framework represents a constraint regarding the assessment of the dataset's model-agnostic robustness. Model architecture selection can significantly influence detection performance, particularly regarding feature extraction capabilities and sensitivity to the specific visual characteristics of construction site imagery. Consequently, the performance metrics reported here should be viewed as a foundational baseline rather than an exhaustive evaluation. Future research efforts are encouraged to validate this dataset against diverse detection architectures, such as Faster R-CNN, DETR, or other state-of-the-art models, to further explore the impact of model choice on detection precision and to establish a broader comparative benchmark.

## Annotation Quality Assessment

The quality of the dataset's annotations was carefully evaluated to ensure reliability and consistency. Inter-annotator agreement was measured using the IoU metric, with a threshold of 0.5 considered acceptable. Multiple annotators labeled overlapping subsets of the dataset, and a senior annotator performed additional quality assurance checks on a selected subset to identify and correct any discrepancies. The results indicated a high level of agreement, with an average IoU of 0.85 and only approximately 3% of annotations requiring revision. This demonstrates that the dataset annotations are accurate and consistent, providing a strong foundation for model training.

## Quantitative Evaluation with Baseline Models

To evaluate the dataset's suitability for model training, we performed quantitative assessments using the baseline object detection model YOLOv11m. Across the initial training epochs, the model achieved a mAP50 of 0.69, a precision of 0.68, a recall of 0.66, and an F1 score of 0.67, demonstrating strong predictive performance, as can be seen in Figure 7.

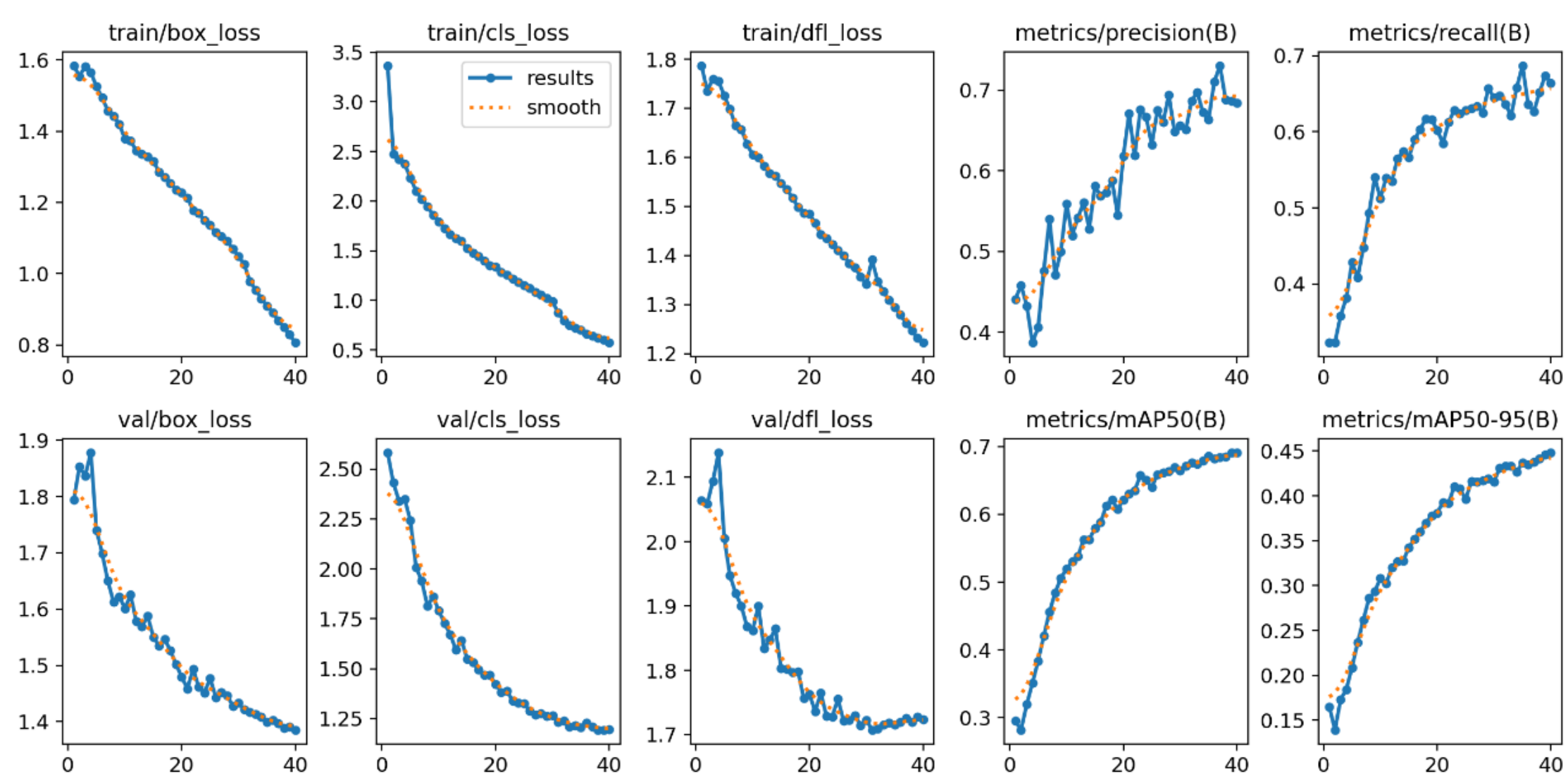

Figure 7. Training loss and validation mAP chart of the models.

To further analyze model performance, class-wise evaluation metrics were computed using the confusion matrix, precision, recall, F1 score, and mAP50. The confusion matrix indicated that the model performed well for most classes, showing strong detection accuracy and limited misclassifications, as can be seen in Figure 8. However, a few classes demonstrated relatively lower performance, likely due to the complex and overlapping nature of construction activities, where visual features are less distinct and more chaotic. These lower-performing classes may benefit from further division into more specific subcategories (level 4) to better capture variations within similar construction tasks. Overall, the results suggest that the model can effectively detect and classify most construction elements, with room for refinement in visually complex categories. While the model demonstrates consistent accuracy across most classes, minor variations were observed, reflecting the influence of class imbalance and intra-class variability on detection performance. A graphical summary of these indicators, including the confusion matrix and class-wise precision, recall, and mAP50, is presented in Figure 8 and Figure 9.

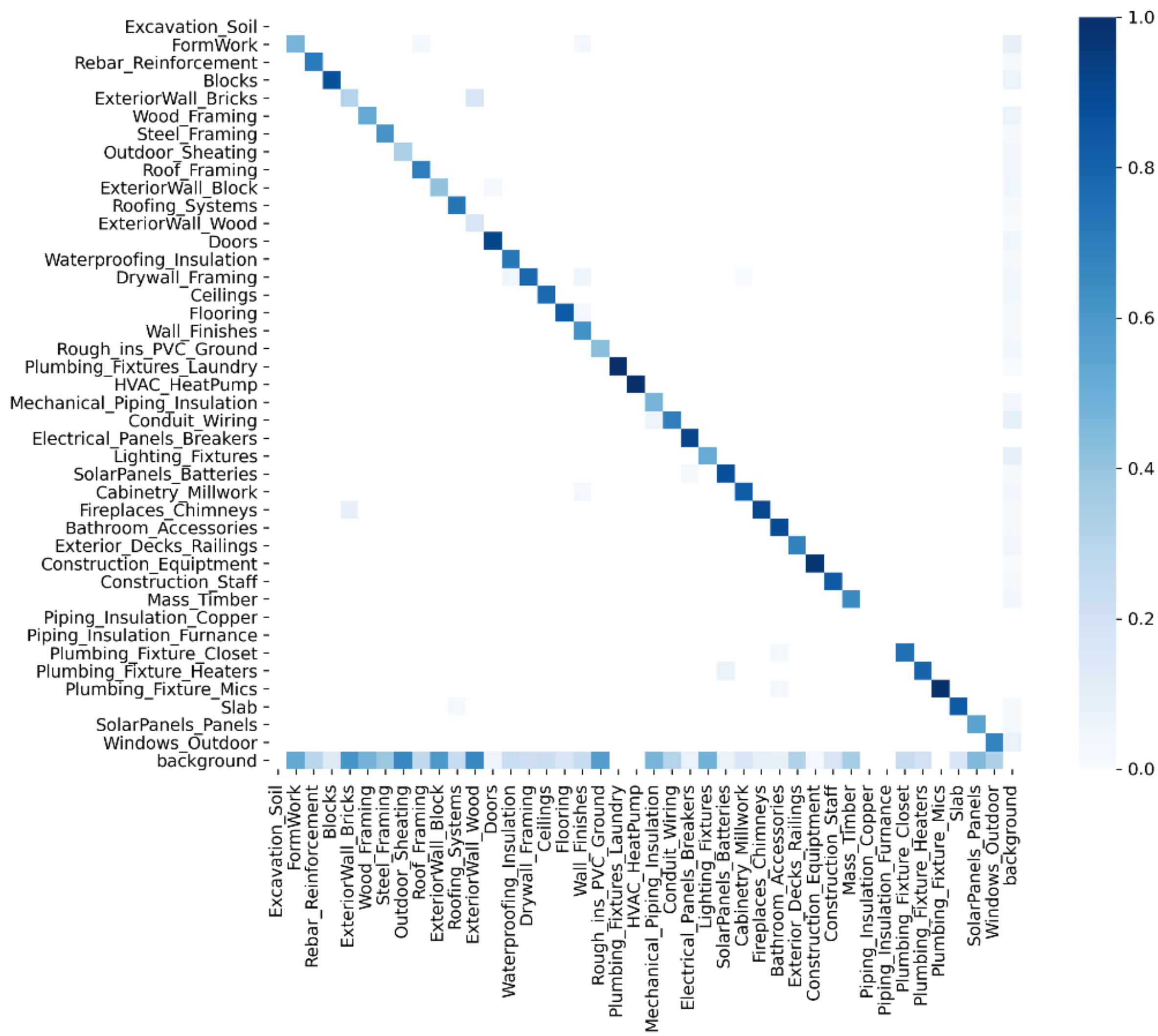

Figure 8. Normalized confusion matrix

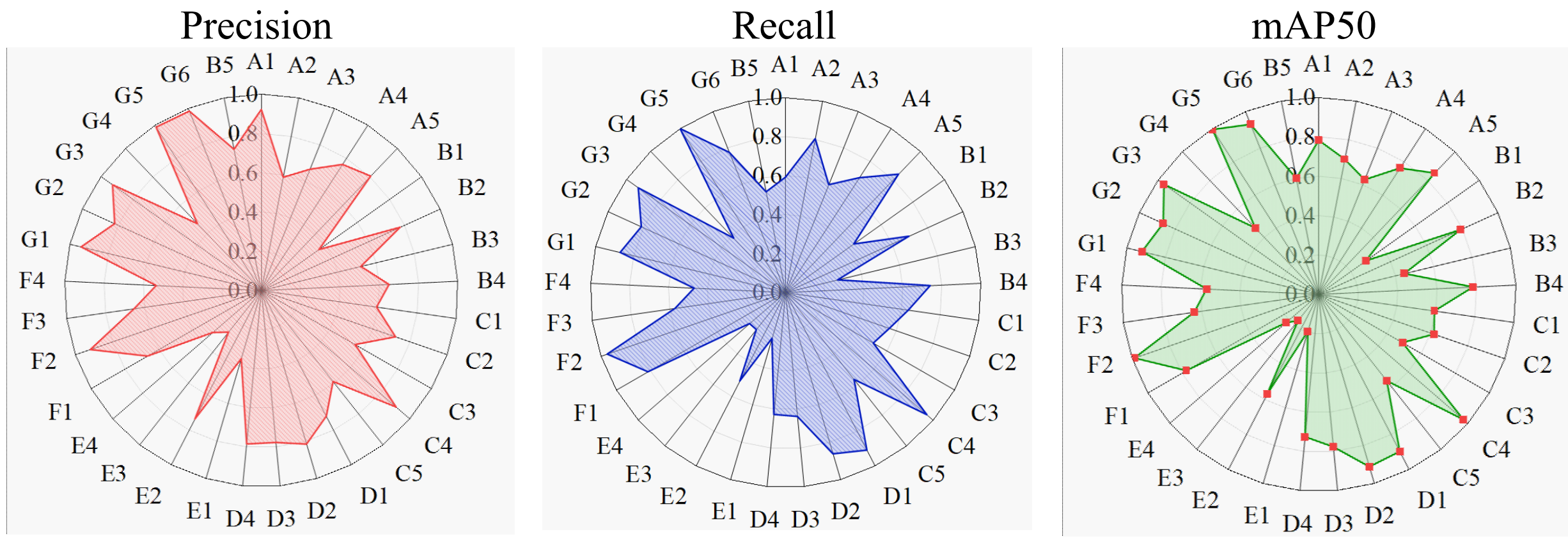

(a) (b) (c)

Figure 9. Class-wise performance: (a) precision, (b) recall, and (c) mAP50.

## Qualitative Analysis

Qualitative analysis was performed through visual inspection of model predictions on sample images from the residential construction divisional dataset, which contains 14 categories. The majority of construction-related objects were accurately detected, demonstrating the reliability of the annotations and confirming the dataset's applicability to real-world residential construction tasks.

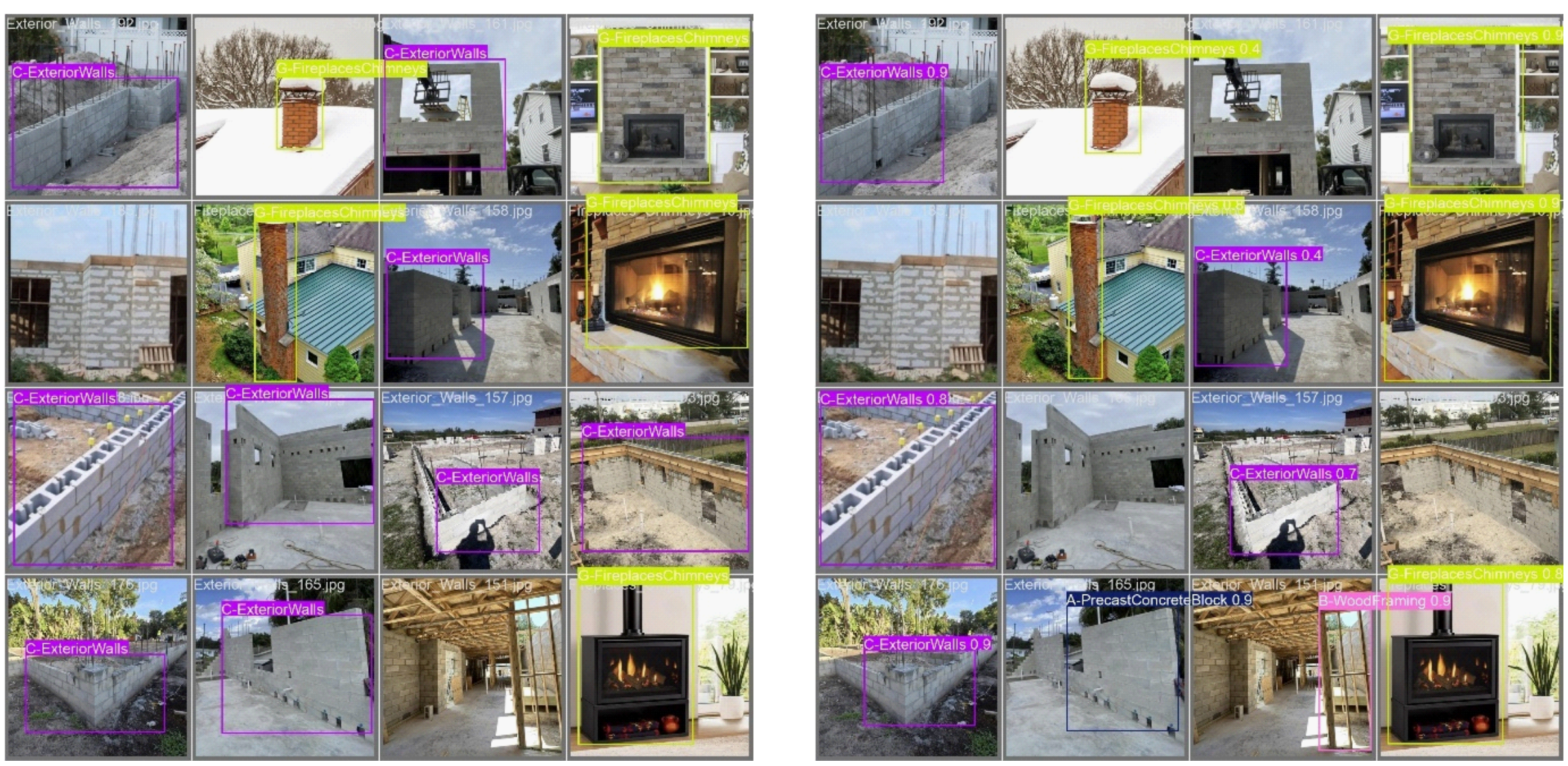

(a) (b)

Figure 10. Qualitative results (a) ground-truth, (b) prediction.

However, challenging cases were observed in images with heavy occlusion, cluttered backgrounds, or extreme lighting conditions, which occasionally led to missed or partial detections. These instances highlight both the complexity of construction sites and the diversity of the dataset.

Example images with annotations were reviewed to demonstrate typical detections, difficult edge cases, and the overall visual complexity, thereby reinforcing the robustness of the dataset. A notable challenge across multiple models was the detection of smaller categories or fine construction details, where bounding boxes were often small, occluded, or visually similar to surrounding materials. This limitation suggests an avenue for future work, particularly in exploring detector architectures better suited for small-object recognition within construction environments. A comparison of qualitative results between ground truth annotations and model predictions is presented in Figure 10.

## Usage Notes

The ResCD-CSI dataset is structured to facilitate broad reuse across diverse deep learning frameworks and research applications. While the repository is organized in a standard YOLO format, it contains high-resolution images paired with independent annotation files that are compatible with any object detection model. This model-agnostic architecture integrates into various pipelines with minimal modification. The dataset is available via figshare[27]. Retaining the original folder hierarchy is recommended to preserve the relational integrity between images and metadata. Labels are stored in TXT files using normalized coordinates. This format is compatible with darknet-based architectures and can be converted to COCO (JSON) or Pascal VOC (XML) for use in frameworks like Detectron2 or TensorFlow. The dataset supports integration into PyTorch or TensorFlow by mapping image paths to label files. This enables deployment across various architectures, including Faster R-CNN, EfficientDet, or YOLO variants. Images are provided in high-resolution JPG format. Users can apply custom resizing, normalization, or augmentation techniques (e.g., mosaic blending) to meet specific model requirements and enhance robustness.

## Data Availability Statement

The data described in this Data Descriptor are available at ResCD-CSI: Residential Construction Image Dataset Using CSI Divisions for Deep Learning-based Object Detection under DOI number 10.6084/m9.figshare.30325087.

## Code availability

All code used for dataset generation, annotation verification, and baseline model evaluation is available in a public GitHub repository (https://github.com/sarbel-org/ResCD-CSI.git). To facilitate clarity and reuse, the repository distinguishes between essential processing scripts and optional baseline models. Core scripts necessary for dataset construction include tools for taxonomic renumbering *'01_Classes_Folder_Name.ipynb'*, batch annotation updates *'02_Annotation_Renumbering.ipynb'*, and automated dataset partitioning *'03_Yolo_Classification.ipynb'* into customized train-val-test ratios. Additionally, an optional training pipeline is provided via a Google Colab-compatible notebook *'04_Construction_Image_Dataset_Yolo11_V1_0.ipynb'* to serve as a baseline implementation using the YOLO framework. The environment utilized Python 3.13, OpenCV, and Pandas, with no proprietary software required for reproduction or extension of the work.

## Acknowledgements

The authors would like to thank and acknowledge Aatith Baswa, Charlotte Smith, Micah Goff, and Iqra Mueed who assisted with dataset annotation. Additionally, we also extend our sincere thanks to Archiwiz for providing access to some of the data, which was essential for the successful completion of this study.

## Funding

This work was conducted under the project "AI-driven Deep Learning for Sustainable Residential Construction" funded by the SUNY ESF Center for AI, Society, and the Environment Faculty Fellowship Program.